\documentclass[11pt,a4paper]{article}
\usepackage[preprint]{acl}
\usepackage{times}
\usepackage[T1]{fontenc}
\usepackage{amsmath,amssymb}
\usepackage{graphicx}
\usepackage{booktabs}
\usepackage{placeins}
\usepackage{url}

\title{MemMend: A Selective Revision Framework\\
for Memory Question Answering}
\author{Sheng Shi \quad Geng Li \\
  Independent Research}
\hypersetup{pdftitle={MemMend: A Selective Revision Framework for Memory Question Answering},
  pdfauthor={Sheng Shi and Geng Li}}

\begin{document}
\maketitle
\raggedbottom

\begin{abstract}
Revising every answer from a long-term memory can consume substantial inference
budget and damage facts that were already correct. We introduce MemMend, a
modular selective revision framework that routes potentially incomplete answers to an
editor, represents corrections as bounded exact-text replacements, and accepts
only edits with resolvable source coordinates and valid edit operations.
The editor preserves the current answer when no legal repair is proposed;
accepted edits require no additional semantic-verification call. On 1,986
LoCoMo development questions, MemMend achieves 67.17 conversation-macro F1 and
88.35 mean Judge score, improving over a verified revision chain by 0.82 and
1.24 points while using 62.37\% fewer revision calls. Routing uses 54.15\%
fewer revision calls than ungated local review, with similar F1 but a 0.67-point
lower Judge score. Applied directly to native PropMem, Mem0, and SimpleMem
answers, the editing interface raises F1 by 3.79, 6.44, and 0.42 points.
It outperforms same-gate rewriting and saves 49.85--83.72\% of editing calls
relative to ungated editing, which retains higher F1 in this adaptation.
A source-by-editing factorial supports local editing for F1 but leaves an
independent quality benefit from source obligations unresolved. Transfer to
451 LongMemEval-V2 Small questions also favors revision efficiency, while an
external PerLTQA dialogue subset reveals a completeness gap. These results
support selective editing as a reusable quality--cost operating point.
\end{abstract}

\section{Introduction}
\label{sec:introduction}

Long-term memory helps agents answer questions about accumulated conversations
and past interactions. Yet retrieving a relevant record does not guarantee that
the resulting answer is complete or correctly scoped. A response may omit one
member of a set, attribute an event to the wrong speaker, or abstain despite
available evidence. Conversely, an answer may already be adequate, and another
unrestricted generation can add unsupported details or overwrite a correct fact.
The decision to revise is therefore part of the memory problem.

Existing memory systems improve storage, retrieval, and consolidation
\citep{packer2023memgpt,xu2025amem,chhikara2025mem0}. Benchmarks such as LoCoMo
and LongMemEval expose the difficulty of using long histories accurately
\citep{maharana2024locomo,wu2025longmemeval}. After retrieval, however, additional
inference still has to be allocated: which answers need review, how much of an
answer may change, and what checks should a proposed change satisfy? Always
rewriting the answer spends calls on easy cases. Independently verifying every
candidate adds another cost, while a source identifier by itself cannot prove
that the cited record entails the proposed claim.

\begin{figure*}[t]
  \centering
  \includegraphics[width=\textwidth]{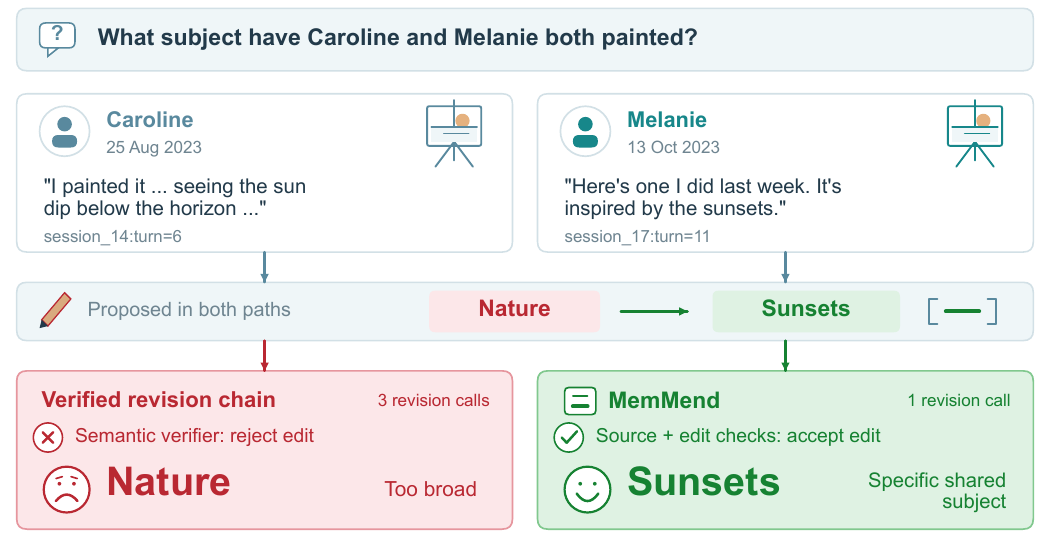}
  \caption{A recorded LoCoMo repair. Both revision paths propose the same
  source-linked replacement, ``Nature'' to ``Sunsets.'' The verified chain
  vetoes the edit and retains the broad answer; MemMend accepts it.
  Answers are verbatim and memory excerpts are shortened. Call counts cover revision.}
  \label{fig:evimem-motivation}
\end{figure*}

MemMend addresses these decisions through a conservative editing policy.
It accepts a memory system's candidate answer and original evidence, allowing
the same revision interface to serve different storage and retrieval backends.
A deterministic gate selects abstentions, list questions, and long answers
for review. The editor proposes either no change or a small set of exact
replacements, accompanied by coordinates into the available original records.
A program checks source resolution, anchor uniqueness, operation permissions,
and issue-specific expansion limits. If these checks fail, the previous
answer is retained after at most one corrective proposal. This design makes
editing inspectable and bounds its scope without paying for a separate
semantic verifier after each edit.

Our evaluation distinguishes this editing interface from the complete upstream
pipeline. Controlled LoCoMo comparisons share candidates and evidence;
an additional study edits native answers from three memory systems.
Figure~\ref{fig:evimem-motivation} traces one recorded repair. We also compare
routing with random allocation, separate source obligations from local editing,
and test transfer to trajectory evidence and external dialogues.

The paper makes three contributions. First, it introduces a selective
revision policy for memory question answering that allocates calls using
observable question and answer properties. Second, it formulates source-linked
local correction as exact edits with deterministic acceptance rules, separating
traceability from semantic support. Third, it provides controlled revision,
routing, and source-by-editing comparisons, including native-answer adaptation
across three memory systems, trajectory transfer, and external dialogues.
These experiments locate the benefits in local repair and call allocation,
while identifying when more extensive revision is needed.

\section{Related Work}
\label{sec:related_work}

\paragraph{Memory organization and evaluation.}
Memory agents organize histories as streams, hierarchical stores, linked
notes, or consolidated facts
\citep{park2023generativeagents,packer2023memgpt,xu2025amem,chhikara2025mem0,liu2026simplemem}.
EviMem combines evidence-gap-driven iterative retrieval with hierarchical
memory organization \citep{li2026evimem}.
LoCoMo evaluates multi-session conversational memory; LongMemEval separates
retrieval, temporal reasoning, updates, and abstention
\citep{maharana2024locomo,wu2025longmemeval}. LongMemEval-V2 adds multimodal
interaction trajectories and questions about states, changes, procedures,
and flawed premises \citep{wu2026longmemevalv2}. PerLTQA supplies synthetic
personal histories with heterogeneous memory types \citep{du2024perltqa}.
MemMend addresses revision after evidence is available. We evaluate its interface
with PropMem from MemEval \citep{prosus2026memeval}, Mem0, and SimpleMem,
keeping each backend's native answer as the starting candidate.

\paragraph{Evidence and grounded generation.}
Retrieval-augmented generation conditions answers on external records
\citep{lewis2020rag}, while retrieval granularity and context position influence
which facts the model uses \citep{chen2024densex,liu2024lostmiddle}.
Self-RAG separates retrieval and support judgments, and ALCE evaluates answer
quality together with citation quality \citep{asai2023selfrag,gao2023alce}.
These distinctions matter for revision: locating a citation is a mechanical
check, whereas deciding that it supports a correction is semantic. MemMend
retains that distinction and tests citation obligations directly, instead of
assuming that traceability guarantees correctness.

\paragraph{Revision as a resource-allocation problem.}
Self-Refine iteratively revises outputs using model-generated feedback
\citep{madaan2023selfrefine}. In memory QA, repeated generation can recover
omitted information but can also replace an adequate answer with an
unnecessary elaboration. We compare iterative revision,
critique followed by rewriting, and a chain with separate semantic verification
under a common additional-call cap. We further compare selective and all-question
editing using shared outputs on their common questions. This isolates the cost
of reviewing more answers from stochastic differences in identical requests.
The generic controls are implemented revision strategies, not reproductions of
complete published memory systems.

\section{Method}
\label{sec:method}

\subsection{Candidate Answers and Original Evidence}
\label{sec:method-evidence}

Let $q$ be a question, $a$ the current answer, and
$\mathcal E=\{(s_i,x_i)\}$ the available evidence. A coordinate $s_i$ identifies
an original conversational turn or a native trajectory span or state; $x_i$ contains
its text and available metadata or images. A source resolver maps a cited
coordinate back to this record. MemMend operates on $(q,a,\mathcal E)$ and can
also use an earlier answer retained by the upstream pipeline. It does not
require reference answers or benchmark category labels.

\begin{figure*}[t]
  \centering
  \includegraphics[width=\textwidth]{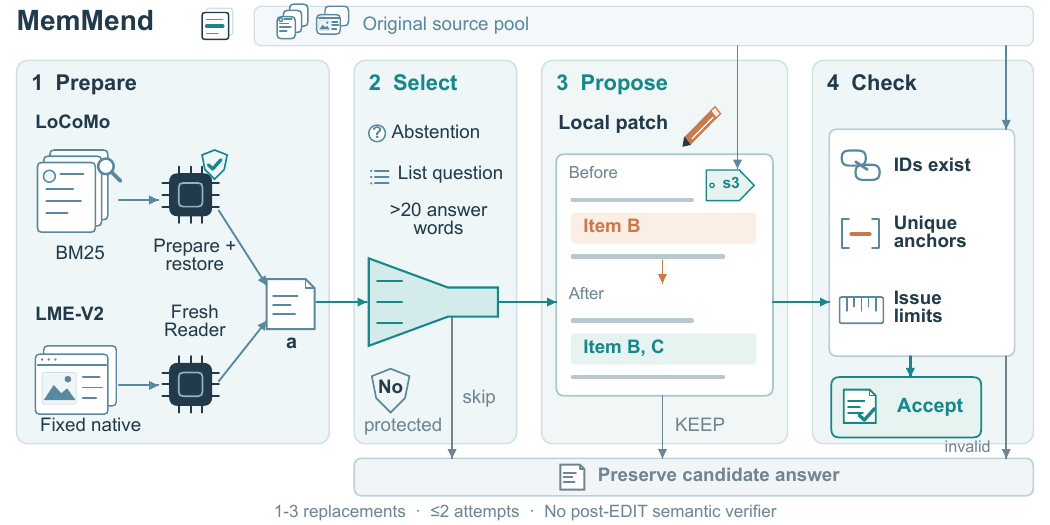}
  \caption{MemMend framework. A selective gate routes candidate answers to
  source-linked local edits. Original memories remain available to the editor
  and deterministic source-coordinate and edit-legality checks; other paths
  preserve the candidate. Preparation supplies either the main conversational
  pipeline's answer, a memory backend's native answer, or a fresh Reader answer
  over fixed trajectory evidence. The highlighted patch is schematic.}
  \label{fig:evimem-framework}
\end{figure*}

Our main conversational configuration retrieves raw sessions with BM25,
selecting five seed sessions and their immediate neighbors within a character
budget. A Raw Session Reader produces an answer, followed by conditional List
Refinement and Scope Checking (SC), then Premise Guard (PG). PG reviews factual
premises and may replace an answer with an abstention when its source and
conflict checks succeed; it does not freely rewrite factual content.
The evaluated main configuration then permits Answer Restoration (AR): if PG
abstains and the saved SC answer is eligible, one independent support check
may restore that exact answer. AR generates no new candidate text.

This upstream configuration is fixed in the main conversational comparisons.
The native-answer adapter instead preserves a backend's answer and resolves
its retrieved memories to their original conversational sources. It applies
the gate, editor, and acceptance checks directly, without SC, PG, or AR.
The LongMemEval-V2 adapter similarly starts from a fresh Reader answer over
fixed trajectory messages and images. All adapters preserve source coordinates;
the upstream stages are configuration choices rather than interface requirements.

\subsection{Selective Routing}
\label{sec:method-routing}

The gate $g(q,a)$ uses inexpensive, deterministic cues. It first protects
explicit negative answers consisting of ``No'' with optional periods,
exclamation marks, or whitespace. Otherwise it
routes an existing abstention, a list-oriented question, or an answer longer
than 20 whitespace-delimited words:
\begin{equation}
 g(q,a)=\neg n(a)\wedge\bigl[u(a)\vee\ell(q)\vee(|a|_w>20)\bigr].
\end{equation}
Here $n$ recognizes the protected negative form, $u$ recognizes explicit
abstention expressions, and $\ell$ applies English surface rules for quantities,
sets, and shared attributes. This gate prioritizes likely omission or scope
problems; it is not a learned uncertainty estimate. Answers with $g=0$ retain
the current candidate without an editor call. Appendix~\ref{app:implementation}
gives the operational rules.

\subsection{Source-Linked Local Corrections}
\label{sec:method-edit}

For a routed answer, the editor returns either \texttt{KEEP} or
\texttt{EDIT}. A correction specifies an issue, one to three anchor--replacement
pairs $\Delta=\{(u_j,v_j)\}$, and one to twelve source coordinates. Each
nonempty anchor $u_j$ must occur exactly once in the original candidate, and
the anchor spans must not overlap. The program constructs $a'=\Delta(a)$ by
replacing spans from right to left. This representation leaves all text outside
the nominated spans unchanged.

Permitted issues include omitted qualifiers or set members, temporal
precision, speaker attribution, incorrect facts, unsupported abstention,
language mismatch, and missing preference reasons. Some require additional
bounds: qualifier, temporal, speaker, and preference-reason repairs may add
at most 6, 8, 4, and 8 distinct normalized words, respectively. Other issue
types retain the anchor and operation checks without these word caps.
A repair of an existing abstention must explicitly identify that issue;
an ordinary edit may not introduce a new refusal. A date already expressed
as an exact calendar day is protected against a temporal-precision edit.
These rules constrain the edit policy, rather than guaranteeing that every
accepted edit is globally minimal.

\subsection{Acceptance and Budget}
\label{sec:method-accept}

Let $V(\Delta,\mathcal E,a)=1$ denote an \texttt{EDIT} whose cited coordinates
resolve, anchors are unique and disjoint, issue and operation are permitted,
and resulting answer satisfies the applicable length bounds. The output is
\begin{equation}
 a_{\mathrm{out}}=
 \begin{cases}
 \Delta(a), & g(q,a)V(\Delta,\mathcal E,a)=1,\\
 a, & \text{otherwise}.
 \end{cases}
\end{equation}
The first valid \texttt{KEEP} or \texttt{EDIT} ends the stage. An invalid
proposal permits one corrective attempt; two invalid proposals preserve the
candidate. No separate semantic verifier follows an accepted edit. Resolving
$s_i$ establishes where the cited evidence comes from, not whether it entails
$v_j$. Semantic support is requested in the editor prompt and tested
empirically, rather than certified by $V$.

The main conversational revision budget is at most three calls: one for AR and
two edit attempts. Unused slots are not reassigned. Native-answer and trajectory
adapters have no AR call and allow two edit attempts. The first valid proposal typically
ends editing early. We report actual calls and tokens, since an identical
cap does not imply identical expenditure.

\section{Experiments}
\label{sec:experiments}

We evaluate whether selective editing improves on the shared
candidate and stronger revision controls; whether routing allocates work
usefully; which effects come from source obligations versus local editing;
and whether the policy transfers across memory backends, evidence formats,
and dialogue sources.

\subsection{Evaluation Protocol}
\label{sec:protocol}

\paragraph{Data and scope.}
The main study uses all 1,986 questions from ten LoCoMo conversations
\citep{maharana2024locomo}. A separate factorial regenerates answers on the
same questions. The trajectory study uses all 451 LongMemEval-V2 Small
questions, comprising 211 Enterprise and 240 Web questions
\citep{wu2026longmemevalv2}. Both collections were previously used for
development; they support controlled mechanism and transfer comparisons,
not independent-source generalization. An external evaluation uses a fixed
200-question PerLTQA dialogue subset from 31 character histories, with no
prior use found in the available local development records. Its inputs
exclude reference-memory pointers and answer anchors.

\paragraph{Models and metrics.}
All new Reader and editor calls use \texttt{gpt-5.6-sol} with low reasoning
and a 2,400-token output cap. LoCoMo uses conversation-macro word-set F1 and
the MemEval Judge rubric \citep{prosus2026memeval}: separate binary accuracy,
relevance, and completeness scores, averaged over questions, with their mean
reported as Judge. Scores are on a 0--100 scale. The Judge is
\texttt{gpt-5.2} with no reasoning; PerLTQA uses the same rubric and
character-macro F1. LongMemEval-V2 uses native final-answer extraction and
native accuracy: 295 rule-graded and 156 model-graded questions, with
\texttt{gpt-5.2} at medium reasoning for the latter. These Judge settings
and metrics are distinct. Model snapshots and scoring details appear in
Appendix~\ref{app:revision-protocol}.

\begin{table*}[t]
  \centering
  \caption{Matched conversational revision on 1,986 LoCoMo questions. Judge accuracy, relevance, and completeness (Acc., Rel., Comp.) and their mean are percentages. Extra calls count revision work after the shared upstream pipeline. Bold marks the best displayed quality scores.}
  \label{tab:revision-locomo-main}
  \begingroup
  \small
  \setlength{\tabcolsep}{3pt}
  \renewcommand{\arraystretch}{1.10}
  \begin{tabular*}{\textwidth}{@{\extracolsep{\fill}}lrrrrrr@{}}
    \toprule
    Method & F1 & Acc. & Rel. & Comp. & Judge mean & Extra calls \\
    \midrule
    Scope-checked answer & 63.96 & 81.37 & 91.49 & 77.74 & 83.53 & 0 \\
    Premise Guard & 63.40 & 80.31 & 89.68 & 76.94 & 82.31 & 0 \\
    Verified revision chain & 66.35 & 84.54 & 94.66 & 82.12 & 87.11 & 2,493 \\
    Iterative revision & 48.22 & 75.38 & 87.36 & 70.39 & 77.71 & 5,958 \\
    Critique and rewrite & 61.97 & 83.03 & 93.96 & 81.12 & 86.04 & 5,958 \\
    \textbf{MemMend} & \textbf{67.17} & \textbf{85.40} & \textbf{96.12} & \textbf{83.53} & \textbf{88.35} & 938 \\
    \bottomrule
  \end{tabular*}
  \endgroup
\end{table*}

\paragraph{Comparators and uncertainty.}
Within the main and factorial studies, arms share fresh upstream answers and evidence. Controls
include the scope-checked answer, PG, a verified abstention/restoration/editing
chain, three-step iterative revision, and critique followed by rewriting
under the same three-call revision cap. Direct-edit ablations remove AR,
routing, or issue-specific added-word bounds. Identical requests reuse the
same outputs and are charged to each standalone policy. We report successful
logical Reader calls; Judge and historical retrieval costs are excluded.
Paired 95\% intervals use 50,000 whole-conversation bootstrap draws for
LoCoMo and whole-character draws for PerLTQA. LongMemEval-V2 uses question
resampling within each domain. Intervals condition on fixed outputs, are
exploratory, and have no multiplicity adjustment.

\subsection{Conversational Quality and Revision Cost}
\label{sec:locomo-results}

Table~\ref{tab:revision-locomo-main} reports the prespecified main configuration.
MemMend improves over PG by 3.78 F1 points ($[2.92,4.60]$) and 6.04 Judge
points ($[4.86,7.11]$). It also improves over the verified chain by
0.82 F1 ($[0.12,1.56]$) and 1.24 Judge ($[0.69,1.90]$), using 938 rather
than 2,493 revision calls. This is a 62.37\% reduction in revision calls;
including the shared upstream, the reduction is 24.46\% (4,802 versus 6,357).
Both generic revision controls use all 5,958 available additional calls and
have lower aggregate F1 and Judge scores. The observed gain thus comes with
less revision work than these controls, rather than a larger inference budget.

Category results reveal a tradeoff: MemMend improves ordinary-question F1
over PG but reduces category-5 abstention performance. Iterative revision
also has higher ordinary-question Judge despite lower F1
(Appendix~\ref{app:revision-locomo}).

\paragraph{Component deletion.}
Removing independent AR gives 67.20 F1 and 88.37 Judge with 909 revision
calls. AR changes 35 intermediate candidates but only three final answers;
these results do not establish its necessity. The editor still receives the
saved SC answer without AR. We retain the prespecified main row and report
this simplification separately. Removing added-word bounds gives 66.83 F1
and 88.18 Judge; both paired differences have intervals crossing zero.
Appendix Table~\ref{tab:revision-locomo-ablation} reports every deletion and
its uncertainty.

\subsection{Revision across Memory Systems}
\label{sec:external-protocol}

\begin{table}[t]
  \centering
  \caption{Complete-system LoCoMo scores on 1,986 questions. F1 and Judge mean are percentages; protocols are described in Section~\ref{sec:external-protocol}. MemMend has the highest displayed F1 and Judge mean, exceeding PropMem by 5.04 and 7.64 points, respectively.}
  \label{tab:revision-external-native}
  \begingroup
  \footnotesize
  \setlength{\tabcolsep}{3pt}
  \renewcommand{\arraystretch}{1.10}
  \begin{tabular*}{\linewidth}{@{\extracolsep{\fill}}lrr@{}}
    \toprule
    System & F1 & Judge \\
    \midrule
    PropMem & 62.13 & 80.72 \\
    Mem0 & 48.14 & 74.77 \\
    SimpleMem & 46.95 & 76.85 \\
    \midrule
    \textbf{MemMend} & \textbf{67.17} & \textbf{88.35} \\
    \bottomrule
  \end{tabular*}
  \endgroup
\end{table}

\begin{table*}[t]
  \centering
  \caption{Editing the original answers of three memory systems on all 1,986 LoCoMo questions. SP: MemMend selective patching; UP: ungated patching; SR: selective rewriting. Bold marks the highest F1 within each row. Calls are incremental logical editor invocations for each variant. SP improves every native baseline and exceeds SR while using fewer calls than UP.}
  \label{tab:revision-native-editor}
  \begingroup
  \small
  \setlength{\tabcolsep}{3pt}
  \renewcommand{\arraystretch}{1.10}
  \begin{tabular*}{\textwidth}{@{\extracolsep{\fill}}lrrrrrrr@{}}
    \toprule
    & \multicolumn{4}{c}{F1} & \multicolumn{3}{c}{Editing calls} \\
    \cmidrule(lr){2-5}\cmidrule(lr){6-8}
    Memory system & Native & SP & UP & SR & SP & UP & SR \\
    \midrule
    PropMem & 62.13 & 65.92 & \textbf{66.47} & 64.88 & 916 & 1,998 & 910 \\
    Mem0 & 48.14 & 54.58 & \textbf{56.58} & 53.72 & 997 & 1,988 & 996 \\
    SimpleMem & 46.95 & 47.37 & \textbf{48.12} & 46.95 & 324 & 1,990 & 325 \\
    \bottomrule
  \end{tabular*}
  \endgroup
\end{table*}

We evaluate PropMem \citep{prosus2026memeval}, Mem0
\citep{chhikara2025mem0}, and SimpleMem \citep{liu2026simplemem} on the same
1,986 LoCoMo questions. Table~\ref{tab:revision-external-native} compares
their saved native answers with the registered MemMend full configuration.
MemMend's 67.17 F1 and 88.35 Judge exceed PropMem's 62.13 and 80.72;
these system-level results use different upstream representations and budgets.

To test the editing interface, we preserve every native answer and expose the
same resolved source ancestry to MemMend selective patching, ungated patching, and
same-gate rewriting. No conversational preparation or restoration is added.
All four arms are complete for all three backends; this study reports full-cohort
F1, with Judge scoring pending (Appendix~\ref{app:revision-backends}).

Selective patching improves native F1 by 3.79, 6.44, and 0.42 points,
respectively; all three paired 95\% intervals are above zero
(Table~\ref{tab:revision-native-editor}). It also exceeds same-gate rewriting
by 1.04, 0.85, and 0.42 points. Ungated patching has higher F1, whereas
selection saves 54.15\%, 49.85\%, and 83.72\% of editing calls.
Thus the interface transfers a quality--cost tradeoff across these backends.
Native-to-edited gains include rereading original sources; the patch-versus-rewrite
control holds source access and routing fixed.

\subsection{Does Routing Allocate Work Usefully?}
\label{sec:routing-results}

Ungated local review retains the operation guards and the option to keep an
answer, but bypasses the selective gate. It uses 2,046 revision calls and scores 67.09 F1 and
89.02 Judge. MemMend saves 54.15\% of these revision calls, or 18.75\%
including the upstream pipeline. The F1 difference is only $+0.08$ points,
with an interval crossing zero; Judge decreases by 0.67 points
($[-1.09,-0.24]$), mainly through completeness. Selection is therefore a
lower-cost operating point, not a quality improvement over ungated review.

To distinguish useful selection from simply reviewing fewer questions, we
reuse the same per-question editing outputs and compare random routing at
matched expected expenditure. Twenty fixed seeds and seven nested coverage
levels define the random policies; mixture weights depend only on recorded
costs. At MemMend's expected call budget, random routing scores 65.21 F1 and
85.59 Judge. MemMend gains 1.96 F1 ($[1.50,2.42]$) and 2.76 Judge
($[2.17,3.25]$); matching expected tokens gives similar results
(Appendix Table~\ref{tab:revision-routing}). These are retrospective
fixed-output comparisons at an expected budget, rather than fresh trials
with identical realized expenditure. They nevertheless show that the gate
selects more useful edits than random allocation in this cohort.

\paragraph{Gate sensitivity.}
Fixed-output branch deletions identify the main routing signal: removing
abstention routing reduces F1 by 2.52 points and Judge by 3.79; removing list
routing reduces them by 0.30 and 0.57. Length routing has much smaller effects
with intervals crossing zero. Alternative length thresholds yield similar
scores at different call counts. The original gate remains the main policy;
all nine policies and paired intervals appear in Appendix~\ref{app:revision-gate}.

\subsection{Source Obligations Versus Local Editing}
\label{sec:factorial-results}

\begin{table}[t]
  \centering
  \caption{Fresh LoCoMo source/locality factorial. Source denotes the EDIT citation obligation; all policies retain the same original evidence. Local edits have higher F1 than whole-answer rewrites.}
  \label{tab:revision-factorial}
  \begingroup
  \small
  \setlength{\tabcolsep}{3pt}
  \renewcommand{\arraystretch}{1.10}
  \begin{tabular*}{\linewidth}{@{\extracolsep{\fill}}llrr@{}}
    \toprule
    Policy & Source & F1 & Judge \\
    \midrule
    Scope-checked answer & -- & 63.79 & 83.28 \\
    Premise Guard & -- & 62.99 & 81.67 \\
    Local edit & Required & 66.86 & 88.03 \\
    Local edit & Not required & 67.00 & 88.12 \\
    Whole-answer rewrite & Required & 66.00 & 88.25 \\
    Whole-answer rewrite & Not required & 66.22 & 88.42 \\
    \bottomrule
  \end{tabular*}
  \endgroup
\end{table}

A separate fresh generation crosses two editor choices: required source
coordinates versus no citation obligation, and local patches versus
whole-answer rewriting. The four arms share their upstream answer,
restoration, evidence, and gate; each routes the same 886 questions.
Removing the citation obligation leaves the original evidence available.
The locality intervention jointly changes repair instructions, exact anchors,
and expansion bounds. Table~\ref{tab:revision-factorial} reports this cohort
separately from the main experiment.

Local editing improves F1 by 0.819 points on average over rewriting; its
Judge effect is unresolved. Requiring coordinates has a $-0.183$-point
marginal F1 effect and no clear Judge benefit. A separate blinded model audit
on 250 preselected questions finds a $-0.205$-point source effect on
unsupported additions, with an interval of $[-0.890,0.438]$.
These results support local editing for lexical overlap, while leaving
independent support gains from source obligations unestablished. Source
coordinates remain useful for locating and inspecting proposed repairs.
Appendix~\ref{app:revision-factorial} reports all factorial contrasts,
interactions, and the audit protocol.

\subsection{Transfer to Native Trajectory Evidence}
\label{sec:lme-results}

\begin{table*}[t]
  \centering
  \caption{LongMemEval-V2 Small transfer on the same 451 fixed evidence packets. Tokens per question include the fresh Reader and revision. Selective patching uses far fewer revision calls than ungated local review, with similar observed accuracy.}
  \label{tab:revision-lme-transfer}
  \begingroup
  \small
  \setlength{\tabcolsep}{3pt}
  \renewcommand{\arraystretch}{1.10}
  \begin{tabular*}{\textwidth}{@{\extracolsep{\fill}}lrrrr@{}}
    \toprule
    Method & Accuracy (\%) & Correct & Extra calls & Tokens/question \\
    \midrule
    Reader & 40.58 & 183/451 & 0 & 14,561 \\
    \textbf{MemMend (selective patching)} & \textbf{41.24} & \textbf{186/451} & 57 & 16,359 \\
    Ungated local review & 41.02 & 185/451 & 451 & 29,916 \\
    Selective rewriting & 40.80 & 184/451 & 58 & 16,399 \\
    Generic revision & 33.70 & 152/451 & 451 & 29,254 \\
    Repeated generic revision & 32.82 & 148/451 & 902 & 43,946 \\
    Critique and rewrite & 36.81 & 166/451 & 902 & 43,942 \\
    \bottomrule
  \end{tabular*}
  \endgroup
\end{table*}

The LongMemEval-V2 study holds each native evidence packet fixed and generates
a new common Reader answer. It tests transfer of the revision policy across
text and image evidence, rather than performance of a newly run retrieval
system. All 451 questions remain in the denominator, including thirteen
with empty memory packets. The shared gate routes 57 questions.

MemMend answers 186 questions correctly (41.24\%), compared with 183 for the
Reader and 185 for ungated local review
(Table~\ref{tab:revision-lme-transfer}). The Reader comparison changes three
normalized answers, all repairs. Accuracy differences versus the Reader,
ungated local review, and selective rewriting are only $+0.67$, $+0.22$,
and $+0.44$ points, respectively; their intervals touch zero.
The stronger result concerns resource use: 57 versus 451 revision calls
saves 87.36\% of revision calls. Including the common Reader yields
508 versus 902 calls and 7.38 versus 13.49 million input-plus-output tokens,
reducing calls by 43.68\% and tokens by 45.32\%.

Generic revision and critique/rewrite reach 33.70\% and 36.81\% accuracy,
respectively. Generic revision repairs three answers but damages 34 that
were previously correct, whereas selective patching makes no observed
wrongward changes. At the selective policy's expected budget, mixtures of
the Reader with generic revision or critique/rewrite also score lower:
by 1.53 and 0.90 accuracy points for call matching. These comparisons reuse
fixed outputs and cost-only mixture weights
(Appendix~\ref{app:revision-lme}). All policies fail on the thirteen
empty-memory packets, emphasizing that answer revision cannot replace
missing evidence.

\subsection{External Dialogue Boundary}
\label{sec:external-results}

On the fixed PerLTQA dialogue subset, MemMend scores 20.65 F1 and 46.83 Judge,
above PG's 18.54 and 41.50. Iterative revision is stronger at 25.92 F1 and
51.50 Judge, including higher completeness (30.50 versus 22.00).
MemMend uses 133 revision calls versus 600, or 569 versus 1,036 calls
including the shared upstream. This external result supports a cheaper
operating point but shows that restrictive editing can leave substantial
information unrecovered. Appendix~\ref{app:revision-perltqa} reports all
arms and the subset construction. An additional coverage-planning attempt
also failed to improve the selected policy and is retained in the appendix.

\section{Conclusion}
\label{sec:conclusion}

MemMend allocates revision selectively and expresses corrections as
source-linked exact edits. On matched LoCoMo comparisons, it improves over
a verified revision chain while reducing calls. Routing preserves most
observed quality at lower cost, and the policy transfers to fixed native
trajectory evidence with few substantive changes. Applied to native PropMem,
Mem0, and SimpleMem answers, the same editing interface improves F1 and
outperforms same-gate rewriting, while ungated editing retains higher F1 at
greater call cost. Factorial controls favor
local editing for F1 but do not establish independent quality gains from
citation obligations. The external dialogue result identifies completeness
as a remaining obstacle. These findings motivate memory systems that choose
when to preserve, locally repair, or more broadly regenerate an answer.

\section*{Limitations}
The decision to revise still has substantial room for improvement. Surface
rules can miss short but incorrect answers and over-select long yet adequate
ones; calibrated, task-adaptive routing could better balance omission risk
and computation. Exact-text patches restrict drift but can be too rigid
when an answer requires reorganization or synthesis across many records.
Future work should choose edit scope adaptively while preserving already
supported content. Source resolution also leaves semantic support unresolved:
a valid coordinate cannot certify speaker identity, temporal scope, or
cross-modal entailment. More reliable support assessment and better recovery
of missing evidence remain necessary, especially when text and images conflict.
Finally, explicit abstentions and negative answers require finer distinctions
between missing information and a contradicted premise, so that preserving
an answer does not also preserve an avoidable error.

\section*{AI Assistance Disclosure}
We used an LLM to assist with drafting and editing manuscript text, literature
search, and figure and table preparation. The authors remain responsible for
the final text, citations, methods, and reported results.

\FloatBarrier
\bibliography{related_work}

\appendix
\section{Implementation Details}
\label{app:implementation}

\paragraph{Evidence adapters.}
Conversational retrieval uses SQLite FTS5 Porter BM25 over raw session text.
Five seed sessions are expanded with immediately adjacent sessions, retaining
whole sessions within a 240,000-character input budget. Retrieval does not
use reference answers, extracted answer propositions, or benchmark labels.
List Refinement and Scope Checking run conditionally on list-routed questions;
other questions retain the preceding answer at these stages.
Session/turn coordinates and dialogue identifiers resolve to original speaker,
text, and timestamp records. The trajectory adapter instead retains native
trajectory/state text and image coordinates; retrieval-agent analysis may
provide context but is not treated as a citable original record.

\paragraph{Gate and proposal contract.}
The protected negative pattern matches case-insensitive ``No'' followed
only by whitespace, periods, or exclamation marks. Abstention patterns include \texttt{None},
\texttt{UNKNOWN}, ``not stated,'' ``not mentioned,'' ``not specified,''
``not available,'' ``insufficient information/evidence,'' and
``no evidence/information.'' List routing uses surface expressions such as
``how many,'' ``what are,'' shared/both/in-common constructions, and selected
plural forms following ``what'' or ``which.'' These rules use the question
and current answer, never the target or a question category.

A valid \texttt{KEEP} has a nonempty reason and requires no source coordinates.
An \texttt{EDIT} has one to three nonempty original anchors, each unique and
changed, with disjoint spans and one to twelve valid source coordinates.
Aliases are normalized and duplicate coordinates removed. The constructed
answer must be nonempty and at most 8,000 characters. Added-word caps count
distinct lowercase regex words newly introduced relative to the current
answer: 6 for a missing qualifier, 8 for temporal precision, 4 for speaker
attribution, and 8 for a preference reason. Set completion, wrong-fact repair,
unsupported abstention, and language correction have no such word cap, while
retaining other edit checks. The original candidate remains the anchor text
for a corrective proposal. Failed network requests do not count as successful
\texttt{KEEP} decisions.

\paragraph{Upstream restoration and verification controls.}
PG skips existing abstentions, explicit negative answers, and hypothetical
questions. Its factual-premise review can retain the answer, request a local
evidence lookup, or propose abstention subject to source, quote, speaker, and
conflict checks. AR nominates the exact saved SC answer only when the current
PG candidate is \texttt{None} and differs from SC, and the saved SC answer
is factually eligible for review. It restores SC only after a successful independent support check.
The no-AR ablation removes that call while retaining SC in editor context.
The verified-chain comparator instead includes Verified Abstention Review
(VAR), AR, and Verified Completion and Recovery (VCR), whose proposed changes
receive semantic verification within the common revision-call cap. The
selected direct editor removes VAR and post-edit VCR verification; it does
not remove the upstream AR verifier from the registered main configuration.

\section{Additional Revision Evaluation}
\label{app:revision-evaluation}

\subsection{Protocols and Cost Accounting}
\label{app:revision-protocol}

The conversational experiments use the \texttt{gpt-5.6-sol} Reader
(\texttt{gpt-5.6-sol-2026-07-09}, low reasoning) with an output limit of
2,400 tokens per call. In the main conversational study, each revision policy
has at most three additional Reader calls after the shared upstream pipeline. Direct editing allocates
at most one call to Answer Restoration (AR) and two proposal attempts;
unused slots are not reassigned. Iterative revision uses three successive
revision calls, while critique and rewrite uses the same call cap. The
verified revision chain applies Verified Abstention Review (VAR), AR, and
Verified Completion and Recovery (VCR), including semantic verification.
These are implemented controls with shared evidence and budgets.

Fresh conversational upstream answers use question-based BM25 session
retrieval followed by the Raw Session Reader, List Refinement, Scope
Checking, and Premise Guard (PG). Within each experiment, policies share
upstream answers and original evidence. The direct-revision study and the
source/locality factorial generate separate upstream answers on the same
1,986 LoCoMo questions; their scores are therefore reported separately.
The routing study reuses the direct-revision study's fixed AR candidates and
ungated editing outputs. The blinded audit samples the factorial's
outputs. A separate coverage-planning calibration generates another matched
output cohort on 150 questions. All ten LoCoMo conversations were previously
used for development.

Conversational F1 is word-set overlap, averaged within each source group and
then equally across groups. The \texttt{gpt-5.2-2025-12-11} Judge separately
scores accuracy, relevance, and completeness; Judge mean averages these
three question-level scores. Conversational scoring uses no reasoning and a
512-token output limit, with the fixed recognized-refusal rule for empty
references. This common rubric is distinct from PerLTQA's official metrics.
Paired intervals use 50,000 whole-source-group bootstrap draws. LoCoMo has
ten groups and the external PerLTQA subset has 31. Intervals condition on
the generated answers and are exploratory, without multiplicity correction.

Extra-call counts include successful logical revision work, including AR
when present. Shared stages are charged to each standalone policy but
executed only once when identical requests can be reused. The common
LoCoMo upstream uses 3,864 calls in the direct experiment and 3,879 in the
factorial; the PerLTQA upstream uses 436. These common-pipeline counts are
not the minimal deployment cost of Scope Checking alone. Token totals sum
input and output tokens, with reasoning tokens already included in output.
They measure algorithmic resource use, not prices or end-to-end latency.

\subsection{Conversational Components and Routing}
\label{app:revision-locomo}

\begin{table*}[t]
  \centering
  \caption{LoCoMo component deletions from the same generated cohort as Table~\ref{tab:revision-locomo-main}. Differences are MemMend minus the row, with paired 95\% intervals in percentage points.}
  \label{tab:revision-locomo-ablation}
  \begingroup
  \footnotesize
  \setlength{\tabcolsep}{3pt}
  \renewcommand{\arraystretch}{1.10}
  \begin{tabular*}{\textwidth}{@{\extracolsep{\fill}}lrrrll@{}}
    \toprule
    Method & F1 & Judge & Extra calls & $\Delta$F1 [95\% CI] & $\Delta$Judge [95\% CI] \\
    \midrule
    MemMend & 67.17 & 88.35 & 938 & -- & -- \\
    w/o restoration & 67.20 & 88.37 & 909 & $-0.03\,[-0.09,+0.00]$ & $-0.02\,[-0.05,+0.00]$ \\
    Ungated local review & 67.09 & 89.02 & 2,046 & $+0.08\,[-0.36,+0.53]$ & $-0.67\,[-1.09,-0.24]$ \\
    w/o edit bounds & 66.83 & 88.18 & 937 & $+0.34\,[-0.03,+0.74]$ & $+0.17\,[-0.19,+0.56]$ \\
    \bottomrule
  \end{tabular*}
  \endgroup
\end{table*}

\begin{table}[t]
  \centering
  \caption{Routing over fixed LoCoMo answers. Random rows average all twenty seed-specific mixtures; their budgets are matched in expectation.}
  \label{tab:revision-routing}
  \begingroup
  \footnotesize
  \setlength{\tabcolsep}{3pt}
  \renewcommand{\arraystretch}{1.10}
  \begin{tabular*}{\linewidth}{@{\extracolsep{\fill}}lrrr@{}}
    \toprule
    Routing policy & F1 & Judge & Extra calls \\
    \midrule
    Restoration only & 63.80 & 83.01 & 60.0 \\
    MemMend routing & 67.17 & 88.35 & 938.0 \\
    Ungated local review & 67.09 & 89.02 & 2046.0 \\
    Random: matched calls & 65.21 & 85.59 & 938.0 \\
    Random: matched tokens & 65.21 & 85.60 & 940.7 \\
    \bottomrule
  \end{tabular*}
  \endgroup
\end{table}

MemMend improves over PG by 3.78 F1 points ($[2.92,4.60]$) and 6.04 Judge
points ($[4.86,7.11]$). Against the verified chain, the gains are 0.82
($[0.12,1.56]$) and 1.24 ($[0.69,1.90]$), using 938 rather than 2,493
additional calls. Including upstream work gives 4,802 versus 6,357 calls,
a 24.46\% reduction. Gains over iterative revision are 18.95 F1 and 10.64
Judge points; gains over critique and rewrite are 5.20 and 2.32 points.
Their respective F1 intervals are $[17.65,20.32]$ and $[4.32,6.10]$;
Judge intervals are $[9.41,11.97]$ and $[1.63,3.05]$.

Table~\ref{tab:revision-locomo-ablation} shows which components contribute.
Removing independent restoration slightly raises both point estimates and
saves 29 calls; the editor still sees the previous Scope Checking answer.
Thus these results do not establish a separate benefit from AR. Removing
routing preserves the other operation guards and increases additional calls
from 938 to 2,046. Its F1 is similar, while Judge mean is 0.67 points higher.
Added-word bounds have small positive point estimates with intervals
crossing zero. The registered MemMend row remains fixed rather than being
replaced by the best deletion after evaluation.

The gains do not come solely from refusal credit. On the 1,540 category
1--4 questions, F1 rises from PG's 54.07 to 59.97 and Judge mean from 77.81
to 86.19. However, iterative revision and critique/rewrite score 88.25 and
86.58 on this subset's Judge mean, despite lower F1. On category 5,
MemMend scores 92.28 F1 and 95.81 Judge, below PG's 95.77 and 97.83.
Category 5 contains 446 questions, including two nonempty references;
it is not identical to the 444-question empty-reference subset.

Table~\ref{tab:revision-routing} evaluates selection conditional on the same
per-question editing outcomes. Unselected questions retain AR's answer.
Twenty fixed random seeds and seven nested coverage levels define random
policies; all are retained. Adjacent coverage levels are mixed using only
recorded costs. At MemMend's expected call budget, random routing reaches
65.21 F1 and 85.59 Judge, giving MemMend gains of 1.96
($[1.50,2.42]$) and 2.76 ($[2.17,3.25]$) points. Matching expected tokens
gives 1.96 ($[1.50,2.41]$) and 2.76 ($[2.16,3.24]$) points.
These are retrospective expected-cost policies, not fresh runs with
identical realized expenditure. Bootstrap draws keep masks and mixture
weights fixed. Ungated review still has the higher Judge point
estimate, showing the quality--cost tradeoff.

\subsection{Source Obligation and Local Editing}
\label{app:revision-factorial}

\begin{table*}[t]
  \centering
  \caption{All source/locality factorial contrasts. Differences and paired 95\% intervals are percentage points. Positive source effects favor required citations; positive locality effects favor local edits.}
  \label{tab:revision-factorial-effects}
  \begingroup
  \footnotesize
  \setlength{\tabcolsep}{3pt}
  \renewcommand{\arraystretch}{1.10}
  \begin{tabular*}{\textwidth}{@{\extracolsep{\fill}}lll@{}}
    \toprule
    Contrast & $\Delta$F1 [95\% CI] & $\Delta$Judge [95\% CI] \\
    \midrule
    Source effect: local edit & $-0.144\,[-0.292,-0.001]$ & $-0.084\,[-0.350,+0.197]$ \\
    Source effect: rewrite & $-0.223\,[-0.653,+0.136]$ & $-0.168\,[-0.394,+0.041]$ \\
    Locality effect: source required & $+0.859\,[+0.357,+1.427]$ & $-0.218\,[-0.503,+0.072]$ \\
    Locality effect: no source obligation & $+0.780\,[+0.342,+1.185]$ & $-0.302\,[-0.596,+0.019]$ \\
    Average source effect & $-0.183\,[-0.386,-0.019]$ & $-0.126\,[-0.343,+0.067]$ \\
    Average locality effect & $+0.819\,[+0.391,+1.242]$ & $-0.260\,[-0.498,+0.028]$ \\
    Source--locality interaction & $+0.079\,[-0.363,+0.574]$ & $+0.084\,[-0.186,+0.368]$ \\
    \bottomrule
  \end{tabular*}
  \endgroup
\end{table*}

\begin{table*}[t]
  \centering
  \caption{Blinded model audit of 250 questions from the fresh LoCoMo factorial. Changed counts are unweighted; percentage rates use inclusion weights and all sampled questions. Repairs/harms are changed answers crossing the audit adequacy boundary.}
  \label{tab:revision-blind-audit}
  \begingroup
  \footnotesize
  \setlength{\tabcolsep}{3pt}
  \renewcommand{\arraystretch}{1.10}
  \begin{tabular*}{\textwidth}{@{\extracolsep{\fill}}llrrrrr@{}}
    \toprule
    Policy & Source & Changed & Adequate & Unsupported additions & Repairs & Harms \\
    \midrule
    Local edit & Required & 16 & 55.97 & 1.23 & 2.02 & 0.00 \\
    Local edit & Not required & 22 & 56.87 & 1.72 & 2.91 & 0.00 \\
    Whole-answer rewrite & Required & 24 & 57.18 & 1.79 & 2.82 & 0.00 \\
    Whole-answer rewrite & Not required & 20 & 56.86 & 1.72 & 2.50 & 0.00 \\
    \bottomrule
  \end{tabular*}
  \endgroup
\end{table*}

The factorial crosses required source coordinates with local editing versus
whole-answer rewriting. Removing the source obligation removes the EDIT
prompt's citation requirement and coordinate validation; original records
remain visible and shared AR remains source-grounded. Local editing bundles
minimal-repair instructions, unique nonoverlapping patch anchors, and
added-word bounds. Rewriting removes those anchors and bounds but retains
the actual-gap, supported-information, refusal, issue, and date guards.
Thus locality tests an editing policy, not output format alone. Each policy
routes the same 886 questions and uses 948--950 additional calls.

Table~\ref{tab:revision-factorial-effects} reports all four simple effects,
both marginal effects, and the source--locality interaction. Local editing
improves F1 by 0.819 points on average, while its Judge interval crosses
zero. Requiring coordinates has a small negative F1 effect and no clear
Judge benefit. The interaction is also unresolved. There is no consistent
token saving from locality across these four policies; routing supplies
the principal evidence for lower revision cost.

On the factorial's 1,542 nonempty-reference questions, source-required local
editing scores 59.682 F1 and 85.884 Judge, above PG's 53.552 and 77.108.
On the 444 empty-reference questions, it scores 91.944 and 95.495,
below PG's 95.955 and 97.523. The locality effect on category 1--4 Judge
is $-0.455$ points ($[-0.803,-0.039]$). These descriptive splits further
separate lexical-overlap gains from completeness and abstention behavior.

The separate blinded model audit fixes 25 questions per conversation,
without consulting answer scores, and uses inverse-inclusion weights.
It shows the actual Reader-visible source pool, the shared AR answer, and
four anonymously ordered final answers. Method names, proposal citations,
reference answers, and benchmark scores are hidden. A response is
audit-adequate when sufficient and either supported or making no factual
claim. These model judgments differ from reference-based accuracy.

Table~\ref{tab:revision-blind-audit} uses all 250 primary questions as the
weighted denominator; unsupported additions count changed answers with at
least one unsupported new claim. The shared AR adequacy rate is 53.96\%.
The marginal source effect on adequacy is $-0.290$ points
($[-0.946,0.348]$), and on unsupported additions is $-0.205$
($[-0.890,0.438]$). Neither establishes an independent support benefit.
The local-edit adequacy effect is $-0.598$ ($[-1.635,0.334]$).
All four policies have zero observed adequate-to-inadequate edits, but
source-required local editing changes none of the 134 initially adequate
sampled answers. Zero observed events do not establish zero editing risk.

The final audit uniformly judges all packets with a strict structured-output
protocol; no earlier partial audit is pooled into the estimates. Its
\texttt{gpt-5.2-2025-12-11} Judge uses no reasoning and a 4,096-token output
limit. Fifty separately fixed reversed-order packets give 90--94\% support
agreement and 94\% sufficiency agreement. These packets are excluded from
the primary estimates. Intervals condition on the sample and one Judge
realization, leaving this observed order sensitivity outside their scope.

\subsection{Trajectory Transfer and Expected Budgets}
\label{app:revision-lme}

\begin{table*}[t]
  \centering
  \caption{Paired trajectory comparisons: selective patching minus each comparator. Wins/losses count questions correct only for selective patching/only for the comparator.}
  \label{tab:revision-lme-contrasts}
  \begingroup
  \footnotesize
  \setlength{\tabcolsep}{3pt}
  \renewcommand{\arraystretch}{1.10}
  \begin{tabular*}{\textwidth}{@{\extracolsep{\fill}}lrl@{}}
    \toprule
    Comparator & Wins/losses & $\Delta$accuracy [95\% CI] \\
    \midrule
    Reader & 3/0 & $+0.67\,[+0.00,+1.55]$ \\
    Ungated local review & 1/0 & $+0.22\,[+0.00,+0.67]$ \\
    Selective rewriting & 2/0 & $+0.44\,[+0.00,+1.11]$ \\
    Generic revision & 35/1 & $+7.54\,[+5.10,+10.20]$ \\
    Repeated generic revision & 39/1 & $+8.43\,[+5.99,+11.09]$ \\
    Critique and rewrite & 25/5 & $+4.43\,[+2.22,+6.87]$ \\
    \bottomrule
  \end{tabular*}
  \endgroup
\end{table*}

\begin{table*}[t]
  \centering
  \caption{Trajectory controls matched to selective patching in expected cost. A mixture revises with probability $p$ and otherwise retains the Reader answer. Differences are selective patching minus the mixture.}
  \label{tab:revision-lme-budget}
  \begingroup
  \footnotesize
  \setlength{\tabcolsep}{3pt}
  \renewcommand{\arraystretch}{1.10}
  \begin{tabular*}{\textwidth}{@{\extracolsep{\fill}}llrrl@{}}
    \toprule
    Revision policy & Matched budget & $p$ & Accuracy (\%) & $\Delta$accuracy [95\% CI] \\
    \midrule
    Generic revision & Calls & 0.1264 & 39.71 & $+1.53\,[+0.87,+2.36]$ \\
    Generic revision & Tokens & 0.1223 & 39.74 & $+1.51\,[+0.85,+2.33]$ \\
    Critique and rewrite & Calls & 0.0632 & 40.34 & $+0.90\,[+0.25,+1.73]$ \\
    Critique and rewrite & Tokens & 0.0612 & 40.35 & $+0.90\,[+0.24,+1.73]$ \\
    \bottomrule
  \end{tabular*}
  \endgroup
\end{table*}

The trajectory study retains all 451 previously exposed Small questions:
211 Enterprise and 240 Web. Each arm receives the same historical evidence
packet, preserving 988 memory-image occurrences and 29 query images.
Thirteen empty-memory packets remain in the denominator. Retrieval-agent
analysis is auxiliary text rather than a citable original source.
The study generates a fresh Reader answer and revision outputs; it does
not regenerate retrieval.

The Reader model and per-call output limit match the conversational runs.
Trajectory revision has at most two additional calls. Selective patching
uses the frozen output-based gate, source-coordinate membership, and
deterministic edit checks; this transfer does not add separate PG, AR, VAR,
or semantic-verifier modules. Ungated local review shares the identical
patch stage where the selective gate admits a question. Selective rewriting
uses the same gate with whole-answer proposals. Generic revision uses one
call; repeated generic revision reports the second call of that same
sequence. Critique and rewrite uses one critique and one rewrite call.

All arms use the native final-answer extraction rule once, followed by the
native benchmark grading protocol. Model-graded questions use
\texttt{gpt-5.2-2025-12-11} with medium reasoning and a 2,048-token output limit. Identical
question/answer/target requests may reuse the exact saved judgment. The
uniform extraction rule is applied before both scoring and measuring
answer changes. Paired intervals use 50,000 question bootstrap draws
stratified by domain. They condition on the fixed questions, evidence,
and outputs; shared histories preclude an independent-source interpretation.

Selective patching makes 57 revision calls versus 451 for ungated local
review, saving 87.36\% of revision calls. Including the Reader, it uses
508 versus 902 calls and 7,377,750 versus 13,492,163 input-plus-output
tokens, a 45.32\% token reduction. Historical retrieval cost is excluded.
Table~\ref{tab:revision-lme-contrasts} shows that accuracy gains over the
Reader, ungated local review, and selective rewriting are only three,
one, and two questions, with intervals touching zero. These results
support revision efficiency more clearly than higher accuracy.

Only three final answers change under selective patching, all from wrong
to correct. Ungated local review changes six answers, correcting three
and damaging one. Generic revision corrects three but damages 34 previously
correct answers; repeated generic revision corrects two and damages 37;
critique and rewrite corrects six and damages 23. All methods score zero
on the thirteen empty-memory cases. Selective patching gains two dynamic
questions and one procedure question over the Reader, with no static or
gotchas gain; generic revision answers one more gotchas question correctly.

Table~\ref{tab:revision-lme-budget} mixes the Reader with each generic
revision family at selective patching's expected total calls or tokens.
The nearest cost-ordered endpoints and mixture weights use costs alone.
Selective patching exceeds generic mixtures by 1.51--1.53 accuracy points
and critique/rewrite mixtures by approximately 0.90 points. These policies
reuse fixed outputs; they do not constitute new realized-budget runs.
Weights stay fixed during bootstrap, which excludes randomized assignment
and model-generation variability.

\subsection{External Dialogue Evaluation}
\label{app:revision-perltqa}

\begin{table*}[t]
  \centering
  \caption{External PerLTQA dialogue subset: 200 questions from 31 characters. Differences are MemMend minus each comparator, with paired 95\% source-group intervals. Iterative revision scores higher than MemMend on both F1 and Judge.}
  \label{tab:revision-perltqa}
  \begingroup
  \footnotesize
  \setlength{\tabcolsep}{3pt}
  \renewcommand{\arraystretch}{1.10}
  \begin{tabular*}{\textwidth}{@{\extracolsep{\fill}}lrrrll@{}}
    \toprule
    Method & F1 & Judge & Extra calls & $\Delta$F1 [95\% CI] & $\Delta$Judge [95\% CI] \\
    \midrule
    Scope-checked answer & 19.89 & 44.17 & 0 & $+0.77\,[+0.32,+1.28]$ & $+2.67\,[+1.15,+4.42]$ \\
    Premise Guard & 18.54 & 41.50 & 0 & $+2.11\,[+1.24,+3.09]$ & $+5.33\,[+3.32,+7.48]$ \\
    Verified revision chain & 20.07 & 44.67 & 286 & $+0.58\,[-0.25,+1.42]$ & $+2.17\,[+0.66,+3.82]$ \\
    Iterative revision & 25.92 & 51.50 & 600 & $-5.27\,[-7.69,-3.09]$ & $-4.67\,[-7.55,-1.81]$ \\
    Critique and rewrite & 21.21 & 45.50 & 600 & $-0.56\,[-2.03,+0.86]$ & $+1.33\,[-1.15,+3.78]$ \\
    MemMend & 20.65 & 46.83 & 133 & -- & -- \\
    \bottomrule
  \end{tabular*}
  \endgroup
\end{table*}

We separately evaluate the English revised dialogue portion of Personal
Long-Term Memory Question Answering (PerLTQA), a synthetic character-memory
dataset \citep{du2024perltqa}.
Deterministic hash selection and character rotation fix 200 questions
covering 31 exactly matched characters, after excluding known question
and source overlaps. The available local development records contained no
earlier PerLTQA use. All character dialogues are available for retrieval;
reference memories, memory anchors, profile summaries, and event summaries
are excluded from inference. This is static complete-memory QA, with no
query-time filtering, and uses the shared F1/Judge rubric rather than the
official full benchmark protocol.

Table~\ref{tab:revision-perltqa} retains the adverse transfer result.
MemMend improves over PG, but iterative revision is higher by 5.27 F1 and
4.67 Judge points. The iterative baseline's completeness is 30.50 versus
MemMend's 22.00. MemMend uses 133 extra calls versus 600 for iterative
revision; including upstream work gives 569 versus 1,036 calls.
This supports a lower-cost operating point, not the best observed external
accuracy. Because these characters were subsequently observed during
development, further questions sharing their histories cannot be treated
as independent new-source confirmation.

\subsection{Coverage-Planning Calibration}
\label{app:revision-coverage}

\begin{table}[t]
  \centering
  \caption{Separate 150-question LoCoMo development calibration of coverage planning. The candidate and all deletions underperform the freshly generated MemMend comparator.}
  \label{tab:revision-coverage}
  \begingroup
  \footnotesize
  \setlength{\tabcolsep}{3pt}
  \renewcommand{\arraystretch}{1.10}
  \begin{tabular*}{\linewidth}{@{\extracolsep{\fill}}lrrr@{}}
    \toprule
    Method & F1 & Judge & Extra calls \\
    \midrule
    Scope-checked answer & 64.13 & 85.56 & 0 \\
    Premise Guard & 63.70 & 84.89 & 0 \\
    Coverage planning & 60.46 & 84.89 & 305 \\
    w/o plan & 62.64 & 86.67 & 155 \\
    w/o editor source text & 63.11 & 88.00 & 305 \\
    w/o restoration & 61.09 & 86.44 & 300 \\
    MemMend & 66.16 & 89.78 & 77 \\
    Iterative revision & 48.17 & 80.22 & 450 \\
    \bottomrule
  \end{tabular*}
  \endgroup
\end{table}

A separate fixed-hash sample takes fifteen questions from each LoCoMo
conversation. Its candidate inserts a question-and-source coverage plan
before minimal editing; the editor receives the resolved original records.
All eight policies in Table~\ref{tab:revision-coverage} are retained.
Coverage planning scores 60.46 F1 and 84.89 Judge versus the freshly
generated MemMend comparator's 66.16 and 89.78, with differences of
$-5.70$ ($[-8.75,-2.47]$) and $-4.89$ ($[-8.00,-1.78]$) points.
It uses 305 extra calls versus 77. All 150 plans are structurally valid,
so the negative result is not caused by planner-format failure.
Deleting the plan or original editor source text improves point estimates
but does not surpass MemMend. The no-restoration comparison also includes
independent editor-generation variation on unchanged candidates, limiting
its component interpretation. Coverage planning was consequently not added
to the selected method or expanded to the full dataset.

\section{Memory-System Adaptation and Gate Sensitivity}
\label{app:revision-followup}

\begin{table*}[t]
  \centering
  \caption{All fifteen native-answer editing contrasts, in F1 points with paired 95\% intervals. SP: selective patching; UP: ungated patching; SR: selective rewriting. Each cell uses all 1,986 questions. SP improves over Native and SR on every backend, while UP attains higher F1 than SP.}
  \label{tab:revision-native-contrasts}
  \begingroup
  \footnotesize
  \setlength{\tabcolsep}{3pt}
  \renewcommand{\arraystretch}{1.10}
  \begin{tabular*}{\textwidth}{@{\extracolsep{\fill}}llll@{}}
    \toprule
    Contrast & PropMem & Mem0 & SimpleMem \\
    \midrule
    SP $-$ Native & $+3.794\,[+2.698,+4.779]$ & $+6.438\,[+5.648,+7.203]$ & $+0.425\,[+0.197,+0.649]$ \\
    UP $-$ Native & $+4.342\,[+3.151,+5.463]$ & $+8.439\,[+7.371,+9.562]$ & $+1.170\,[+0.681,+1.701]$ \\
    SR $-$ Native & $+2.759\,[+1.699,+3.732]$ & $+5.584\,[+4.777,+6.351]$ & $+0.007\,[-0.361,+0.395]$ \\
    SP $-$ UP & $-0.548\,[-1.075,-0.061]$ & $-2.001\,[-2.692,-1.327]$ & $-0.745\,[-1.193,-0.262]$ \\
    SP $-$ SR & $+1.035\,[+0.443,+1.589]$ & $+0.853\,[+0.511,+1.149]$ & $+0.418\,[+0.193,+0.632]$ \\
    \bottomrule
  \end{tabular*}
  \endgroup
\end{table*}

\begin{table}[t]
  \centering
  \caption{All external-system answer paths on 1,986 LoCoMo questions. Source Reader regenerates answers from expanded original sources; the two subsequent paths apply selective MemMend or ungated local review to that Reader. MemMend increases F1 and Judge over Source Reader for all three systems. PropMem retains a higher native F1.}
  \label{tab:revision-external-source}
  \begingroup
  \footnotesize
  \setlength{\tabcolsep}{3pt}
  \renewcommand{\arraystretch}{1.10}
  \begin{tabular*}{\linewidth}{@{\extracolsep{\fill}}llrr@{}}
    \toprule
    Memory system & Answer path & F1 & Judge \\
    \midrule
    PropMem & Native & 62.13 & 80.72 \\
     & Source Reader & 51.06 & 84.09 \\
     & Source + MemMend & 51.64 & 85.16 \\
     & Source + ungated & 51.51 & 85.68 \\
    \midrule
    Mem0 & Native & 48.14 & 74.77 \\
     & Source Reader & 51.95 & 84.94 \\
     & Source + MemMend & 52.65 & 86.15 \\
     & Source + ungated & 52.19 & 86.74 \\
    \midrule
    SimpleMem & Native & 46.95 & 76.85 \\
     & Source Reader & 50.96 & 84.54 \\
     & Source + MemMend & 51.74 & 85.82 \\
     & Source + ungated & 51.53 & 86.39 \\
    \bottomrule
  \end{tabular*}
  \endgroup
\end{table}

\subsection{Backend Protocol and Native-Answer Editing}
\label{app:revision-backends}

The comparison in Section~\ref{sec:external-protocol} uses all 1,986 LoCoMo
questions for PropMem, Mem0, and SimpleMem. Native denotes each backend's
memory construction, retrieval, and answering implementation under the
specified model configuration. PropMem reuses frozen \texttt{gpt-4.1-mini}
memory extraction and cached embeddings, followed by a new Sol Reader.
Its retrieval limits are 30 propositions and three chunks. Its retained extraction
cost is 272 successful extraction calls and
589,657 input-plus-output tokens. Mem0 (\texttt{mem0ai} 1.0.11) and
SimpleMem (0.1.0) construct memories with Sol and an 8,192-token output cap.
Mem0 uses native inferred updates over batches of ten records, local
Qdrant storage, and retrieval of up to twenty memories. SimpleMem uses
forty-record windows and hybrid retrieval with semantic/keyword/structured
limits of 25/5/5, planning enabled, and at most two reflection rounds.
New embeddings use \texttt{text-embedding-3-small} with 1,536 dimensions.
New answer and editor calls use \texttt{gpt-5.6-sol-2026-07-09}, low
reasoning, and a 2,400-token output cap. Thus PropMem's memory construction
has a different model history from the other backends.

The native-answer follow-up fixes its design after the source-campaign
results were inspected and retains every question. It preserves each
backend's exact native answer as both the current and previous candidate.
The editor receives the union of tracked original records associated with
the retrieved memories, in chronological order without truncation.
This ancestry records computational provenance, not an annotation of
entailed facts. Source pools differ across backends but are identical
within each backend's editing comparison. Reference answers, benchmark
categories, evidence annotations, and scores are excluded from generation.

MemMend selective patching (SP) applies the frozen gate and source-required exact
edits. Ungated patching (UP) bypasses the gate while retaining explicit-No
protection and all edit-operation guards. Selective rewriting (SR) shares
SP's gate and original records, but proposes a complete replacement answer.
The patch--rewrite contrast changes the existing locality policy, including
exact anchors and added-word bounds. Each arm permits at most two proposals;
invalid proposals ultimately retain the native answer. No new retrieval,
Scope Checking, Premise Guard, restoration, or post-edit semantic verifier
is added. Native-to-edited gains therefore include rereading original
sources as well as revision.

All 23,832 answer bindings are complete. Every native answer and its
per-question F1 match the source campaign. We report full-cohort F1;
Judge evaluation of the edited native answers remains incomplete.
Table~\ref{tab:revision-native-contrasts} contains all fifteen specified
F1 contrasts. SP improves on Native and SR for each backend, but UP has
higher F1 throughout. SP changes 281 PropMem, 374 Mem0, and 80 SimpleMem
answers. Its EDIT-call reductions relative to UP are 54.15\%, 49.85\%,
and 83.72\%, respectively. Calls are successful logical allocations to
each standalone policy, excluding native construction and querying.
Identical effective SP/UP requests share saved responses on selected
questions; their allocations must not be summed as physical requests.
Paired 95\% percentile intervals use 50,000 whole-conversation draws
(seed 20261115), conditional on saved outputs and without multiplicity
adjustment or model-regeneration variation.

\subsection{Source Expansion with a Common Reader}
\label{app:revision-source-expansion}

The earlier source-expansion experiment replaces the native answer with
a fresh common Reader answer over the same original-record ancestry used
by both downstream editing arms. Native summaries are omitted from that
Reader's input. Source MemMend and source ungated review then apply selective
and ungated patching to this new candidate, without conversational
restoration stages. All twelve backend--policy combinations have complete
F1 and Judge scores (Table~\ref{tab:revision-external-source}).

Relative to the common Source Reader, selective editing adds 0.59, 0.70,
and 0.79 F1 points for PropMem, Mem0, and SimpleMem, and 1.07, 1.21,
and 1.28 Judge points. Ungated review has higher Judge scores than
selective editing on all three backends. Source expansion itself changes
both representation and Reader behavior: PropMem's F1 drops from 62.13
to 51.06, despite a Judge increase from 80.72 to 84.09. Mem0 and
SimpleMem instead improve on both metrics. These differences motivate
evaluating the native-answer interface separately, rather than assuming
that replacing compressed memories with their sources always helps.
The complete MemMend result of 67.17 F1 and 88.35 Judge retains its original
pipeline and saved outputs; it is excluded from these within-backend
editing contrasts and has a different total inference budget.

\subsection{Fixed-Output Gate Sensitivity}
\label{app:revision-gate-sensitivity}
\label{app:revision-gate}

\makeatletter
\setlength{\@fptop}{0pt}
\setlength{\@dblfptop}{0pt}
\setlength{\@fpsep}{14pt}
\setlength{\@dblfpsep}{14pt}
\makeatother

\begin{table*}[t]
  \centering
  \caption{Gate sensitivity with fixed LoCoMo answers. The primary threshold remains 20 whitespace-separated words; the length branch fires strictly above the threshold. Extra calls include restoration and editing. Among branch deletions, removing abstention has the largest effect; lowering the threshold increases calls for a small quality gain. Bold marks the highest displayed quality scores.}
  \label{tab:revision-gate-sensitivity}
  \begingroup
  \small
  \setlength{\tabcolsep}{3pt}
  \renewcommand{\arraystretch}{1.10}
  \begin{tabular*}{\textwidth}{@{\extracolsep{\fill}}lrrrrr@{}}
    \toprule
    Routing policy & Length threshold & Selected & F1 & Judge & Extra calls \\
    \midrule
    \textbf{MemMend routing (primary)} & 20 & 877 & 67.17 & 88.35 & 938 \\
    Without abstention branch & 20 & 368 & 64.65 & 84.56 & 429 \\
    Without list branch & 20 & 681 & 66.88 & 87.78 & 741 \\
    Without length branch & -- & 795 & 67.14 & 88.17 & 856 \\
    \midrule
    MemMend routing & 10 & 1,103 & \textbf{67.26} & 88.50 & 1,164 \\
    MemMend routing & 30 & 813 & 67.15 & 88.25 & 874 \\
    MemMend routing & 40 & 801 & 67.14 & 88.18 & 862 \\
    \midrule
    Restoration only & -- & 0 & 63.80 & 83.01 & 60 \\
    Ungated local review & -- & 1,986 & 67.09 & \textbf{89.02} & 2,046 \\
    \bottomrule
  \end{tabular*}
  \endgroup
\end{table*}

\begin{table*}[t]
  \centering
  \caption{Gate sensitivity: all nine policies compared with the primary gate and with restoration only. Differences and paired 95\% intervals are points. The threshold is the strict whitespace-word cutoff for the length branch. Four decimals retain the small effects near zero.}
  \label{tab:revision-gate-contrasts}
  \begingroup
  \footnotesize
  \setlength{\tabcolsep}{3pt}
  \renewcommand{\arraystretch}{1.08}
  \begin{tabular*}{\textwidth}{@{\extracolsep{\fill}}lrll@{}}
    \toprule
    Policy & Threshold & $\Delta$F1 [95\% CI] & $\Delta$Judge [95\% CI] \\
    \midrule
    \multicolumn{4}{l}{\textit{A. Policy minus primary MemMend routing}} \\
    \midrule
    MemMend routing (primary) & 20 & $+0.0000\,[+0.0000,+0.0000]$ & $+0.0000\,[+0.0000,+0.0000]$ \\
    Without abstention branch & 20 & $-2.5220\,[-3.2296,-1.8007]$ & $-3.7932\,[-4.5293,-2.9574]$ \\
    Without list branch & 20 & $-0.2956\,[-0.4472,-0.1611]$ & $-0.5707\,[-0.8645,-0.3044]$ \\
    Without length branch & -- & $-0.0279\,[-0.0612,+0.0046]$ & $-0.1846\,[-0.4264,+0.0317]$ \\
    MemMend routing & 10 & $+0.0942\,[+0.0191,+0.1900]$ & $+0.1511\,[+0.0162,+0.3160]$ \\
    MemMend routing & 30 & $-0.0221\,[-0.0497,+0.0035]$ & $-0.1007\,[-0.2050,+0.0000]$ \\
    MemMend routing & 40 & $-0.0322\,[-0.0653,-0.0003]$ & $-0.1678\,[-0.3456,+0.0000]$ \\
    Restoration only & -- & $-3.3685\,[-4.1461,-2.5992]$ & $-5.3374\,[-6.3109,-4.2235]$ \\
    Ungated local review & -- & $-0.0757\,[-0.5373,+0.3683]$ & $+0.6714\,[+0.2362,+1.0953]$ \\
    \midrule
    \multicolumn{4}{l}{\textit{B. Policy minus restoration only}} \\
    \midrule
    MemMend routing (primary) & 20 & $+3.3685\,[+2.5992,+4.1461]$ & $+5.3374\,[+4.2235,+6.3109]$ \\
    Without abstention branch & 20 & $+0.8465\,[+0.5402,+1.1728]$ & $+1.5441\,[+1.1177,+1.9465]$ \\
    Without list branch & 20 & $+3.0729\,[+2.3162,+3.8067]$ & $+4.7667\,[+3.7415,+5.6812]$ \\
    Without length branch & -- & $+3.3406\,[+2.5792,+4.1165]$ & $+5.1527\,[+4.1736,+6.0395]$ \\
    MemMend routing & 10 & $+3.4627\,[+2.6841,+4.2664]$ & $+5.4884\,[+4.3716,+6.4833]$ \\
    MemMend routing & 30 & $+3.3464\,[+2.5780,+4.1293]$ & $+5.2367\,[+4.1823,+6.1653]$ \\
    MemMend routing & 40 & $+3.3363\,[+2.5728,+4.1130]$ & $+5.1695\,[+4.1502,+6.0797]$ \\
    Restoration only & -- & $+0.0000\,[+0.0000,+0.0000]$ & $+0.0000\,[+0.0000,+0.0000]$ \\
    Ungated local review & -- & $+3.2928\,[+2.2884,+4.3567]$ & $+6.0087\,[+4.7711,+7.2436]$ \\
    \bottomrule
  \end{tabular*}
  \endgroup
\end{table*}

This retrospective analysis retains all ten development conversations and
replays the exact shared AR candidate and ungated patch outcome. No new
Reader or Judge call is made. We delete each of the gate's three OR terms
separately and vary the strict whitespace-word length threshold to 10,
30, or 40. Surviving terms can still select a case after a branch deletion;
their effects are not additive. Every skipped case retains AR's answer,
and explicit-No protection, evidence, prompts, edit bounds, and proposal
limits remain fixed. Ungated review selects all questions, but six
protected negatives consume no EDIT call.

Tables~\ref{tab:revision-gate-sensitivity}
and~\ref{tab:revision-gate-contrasts} retain all nine policies and their
paired F1/Judge contrasts against both the primary gate and restoration
alone. Removing the abstention branch reduces F1/Judge by 2.52/3.79
points; removing the list branch reduces them by 0.30/0.57. Length-branch
deletion has small effects whose intervals cross zero. The threshold of
10 adds 226 calls for 0.094 F1 and 0.151 Judge points; the original
threshold of 20 remains primary. Costs include shared AR and selected
editing, with the common 3,864-call upstream counted once in full-pipeline
totals. The paired bootstrap uses 50,000 whole-conversation draws (seed
20261115), holding outputs and masks fixed. These results
describe sensitivity and call allocation on the existing cohort.

\end{document}